# Tangent Bundle Manifold Learning via Grassmann&Stiefel Eigenmaps

Alexander V. Bernstein[1,2], Alexander P. Kuleshov[1,3]

**Abstract** One of the ultimate goals of Manifold Learning (ML) is to reconstruct an unknown nonlinear low-dimensional manifold embedded in a high-dimensional observation space by a given set of data points from the manifold. We derive a local lower bound for the maximum reconstruction error in a small neighborhood of an arbitrary point. The lower bound is defined in terms of the distance between tangent spaces to the original manifold and the estimated manifold at the considered point and reconstructed point, respectively. We propose an amplification of the ML, called Tangent Bundle ML, in which the proximity not only between the original manifold and its estimator but also between their tangent spaces is required. We present a new algorithm that solves this problem and gives a new solution for the ML also.

## 1 Introduction

The goal of Dimensionality Reduction (DR) is extracting low-dimensional structure from high-dimensional data. There exist a number of methods (techniques) for the DR. Linear DR is well known and uses such techniques as Principal Component Analysis[40] (PCA) and classical metric MultiDimensional Scaling[23] (MDS). Various non-linear DR techniques are based on Neural networks-based approaches[2],[63], Self-organizing Maps[44], Generative Topographic Mapping[14], Topology representing networks[56], Diffusion Maps[48], Kernel PCA[67], and others.

A newly emerging direction in the fields of the DR, which has been a subject of intensive research over the last decade consists in constructing a family of DR-algorithms based on studying the local structure of a given dataset: Locally Linear Embedding, LLE[64]; Laplacian Eigenmaps, LE[3]; Hessian Eigenmaps, HLLE[26]; ISOmetric MAPing, ISOMAP[76]; Maximum Variance Unfolding, MVU[82]; Manifold charting[15]; Local Tangent Space Alignment, LTSA[85 - 86], and others. Some of these algorithms (LLE, LE, ISOMAP, MVU) can be considered in the same framework, based on the Kernel PCA algorithm[67] applied to various data-based kernels[4 - 5], [65 - 66].

The goal of the DR techniques in most of these papers is only solving the **Embedding problem**: based on a given dataset (sample)

$$\mathbf{X}_n = \{X_1, X_2, \ldots, X_n\} \subset \mathbf{X} \subset R^p, \qquad (1)$$

consisting of n points sampled from an unknown **Data space X** lying in Euclidean space $R^p$ of high dimension p, find an **'n-point' Embedding mapping**

$$h_{(n)}: \mathbf{X}_n \subset R^p \to \mathbf{Y}_n = h_{(n)}(\mathbf{X}_n) = \{y_1, y_2, \ldots, y_n\} \subset R^q \qquad (2)$$

of the dataset $\mathbf{X}_n$ to a dataset $\mathbf{Y}_n$ lying in an Euclidean space $R^q$ of low dimension q < p, such that the embedded dataset $\mathbf{Y}_n$ 'faithfully represents' the high-dimensional dataset $\mathbf{X}_n$ while inheriting certain subject-driven data properties like preserving the local data

This work is partially supported by Laboratory for Structural Methods of Data Analysis in Predictive Modeling, MIPT, RF government grant, ag. 11.G34.31.0073.
Corresponding authors: A. Bernstein, Email abernstein@hse.ru; A. Kuleshov, Email kuleshov@iitp.ru.
[1] Department of Technologies of Complex Systems Modeling, National Research University Higher School of Economics, Moscow 109028, Russia.
[2] Institute for System Analysis, Russian Academy of Sciences, Moscow 117312, Russia.
[3] Institute for Information Transmission Problems, Russian Academy of Sciences, Moscow 127994, Russia.

geometry (LLE, LTSA), proximity relations (LE, HLLE), geodesic distances (ISOMAP), angles (Conformal Eigenmaps[68]; Conformal ISOMAP[73], Landmark ISOMAP[73]), etc. The embedded points from the dataset $\mathbf{Y}_n$ can have the various names: the embeddings, the features (in Feature Learning, Pattern Recognition), the representations (in Representation Learning[8]), etc.

**Note.** The DR problems are formulated in various ways, and there is no generally accepted terminology in the DR. Thus, some terms introduced in this paper can be different from those used in some other works.

The term *'faithfully represents'* is not formalized in general, and in various methods it is different due to choosing some 'geometrically motivated' optimized cost function $L_{(n)}(y_1, y_2, \ldots, y_n|\mathbf{X}_n)$ which reflects desired properties of the Embedding $h_{(n)}$ (2). For example, in the classical MDS

$$L_{(n)}(y_1, y_2, \cdots, y_n|\mathbf{X}_n) = \Sigma_{i,j}\big(\rho(X_i, X_j) - \|y_i - y_j\|\big),$$

where $\rho$ is a chosen metric in the Data space $\mathbf{X}$. Note that the MDS and PCA methods are equivalent when $\rho$ is the Euclidean metric in $R^p$. As is pointed out in Ref. [18], a general view on the DR can be based on the concept of cost functions.

Another approach for an answering to the question 'What makes a representation good?' is proposed in Ref. [8]: a quality of the Embedding is defined through the general-purpose properties of the Artificial Intelligence tasks (AI-tasks[7]) in which various Machine Learning methods are applied to the data representations (features).

Based on some solution $h_{(n)}$ for a sample $\mathbf{X}_n$, the problem of constructing the mapping h for new **Out-of-Sample** (OoS) points $X_{new} \in \mathbf{X} / \mathbf{X}_n$ could certainly be solved with regard to the dataset $\{\mathbf{X}_n \cup X_{new}\}$; however, the embedding $\mathbf{Y}_n = h_{(n)}(\mathbf{X}_n)$ obtained previously for the initial sample $\mathbf{X}_n$ will not be preserved in the general case. An '**OoS extension**' of the algorithms LLE, LE, ISOMAP and MVU, which are based on Kernel PCA approach, has been proposed in Ref. [5] with using Nyström's eigendecomposition technique[19], [67]. The Cost functions concept[18] allows also constructing an Embedding mapping for the OoS extension problem, another OoS techniques are proposed in Ref. [29], [75], and others.

Thus, the OoS extension methods are a solution of an **Extended Embedding problem** consisting in constructing a **general Embedding mapping**

$$h : \mathbf{X} \subset R^p \to \mathbf{Y} = h(\mathbf{X}) \subset R^q \qquad (3)$$

of the Data space $\mathbf{X}$ to a **Feature space** $\mathbf{Y} = h(\mathbf{X})$ which preservs specific properties of the Data space. The definition of the Extended Embedding problem uses values of the function h (3) for the OoS points; so, we must define a **Data Model** describing the Data space $\mathbf{X} \subset R^p$, and a **Sampling Model** offering a way for extracting both the sample $\mathbf{X}_n$ and OoS points from the Data space $\mathbf{X}$.

As a rule, the real world data presented in high-dimensional spaces is likely to concentrate in the vicinity of non-linear submanifold of much lower dimensionality[20], [59] (usually called the Manifold hypothesis[59], [62] or the Manifold assumption[54]); a testing this hypothesis is considered in Ref. [59]. Thence, the most popular models in the DR are **Manifold Data Models**[17], [20], [24 - 25], [31], [41], [53 - 55], [78], [85 - 86], in which the Data space is a q-dimensional **Data manifold** (D-manifold) embedded in a p-dimensional space, q < p.

The DR with Manifold Data Models are usually referred to as the **Manifold Learning** (ML), and the DR methods which are motivated by the belief that the Manifold hypothesis is true, are usually referred to as the ML methods[20], [38]. A strict formal definition of the ML will be proposed in Section 2.

The most of the ML methods assume that the D-manifold can be modeled using a single coordinate chart, and only a few methods[15] attempt to deal with multiple charts without assuming a number of the charts known somehow.

In the paper, we will consider only the D-manifold
$$\mathbf{X} = \{X = f(b) \in R^p : b \in \mathbf{B} \subset R^q\} \subset R^p \quad (4)$$
covered by a single coordinate chart f as a Data space. The **Coordinate space B** is an open subset in $R^q$, and f is a diffeomorphism from **B** to **X** = f(**B**) with a differentiable inverse map. It is also assumed that **X** has no self-intersections.

The **Sampling Model** is typically defined as a probability measure μ in the σ-algebra of measurable subsets of the Data space **X** whose support Supp(μ) coincides with **X**. In accordance with this model, the dataset $\mathbf{X}_n$ (1) and the OoS points $X \in \mathbf{X} / \mathbf{X}_n$ are selected from the manifold **X** independently of each other according to the probability measure μ.

The DR problem considered as the general Embedding Problem with the Manifold Data Models can be referred to as the **Manifold Embedding** (ME).

If the Embedding mapping h (3) in the ME preserves only specific properties of high-dimensional data then a substantial data losses are possible when using a reduced q-dimensional vector y = h(X) instead of the initial p-dimensional vector X. As is pointed out in Ref. [50 - 51], one objective of the DR is to preserve as much available information contained in the sample as possible. Consequently, the possibility of reconstructing high-dimensional points X from low-dimensional embeddings h(X) can be a valid evaluation measure for the DR procedures.

To prevent these losses in the general case, the embedding mapping must provide ability for reconstructing the initial vector $X \in \mathbf{X}$ from a vector y = h(X) with small reconstruction error. Thus, it is necessary to construct a **Reconstruction mapping**
$$g: \mathbf{Y}_g \subset R^q \to R^p, \quad (5)$$
defined on the domain **Y** = h(**X**). This mapping determines a **reconstructed value**
$$X^* = (g \cdot h)(X) = g(h(X)) \quad (6)$$
which is the result of successively applying the embedding and reconstruction mappings to a vector $X \in \mathbf{X}$. The Reconstruction mapping g must be defined not only on the Embedded dataset $\mathbf{Y}_n$ (2) (with an obvious reconstruction), but also for 'OoS' embeddings $y = h(X) \in \mathbf{Y} / \mathbf{Y}_n$ obtained by embedding of the OoSpoints $X \in \mathbf{X} / \mathbf{X}_n$.

The Reconstruction mapping g determines a **Reconstruction error**
$$\delta(X) = \|X - g(h(X))\| \quad (7)$$
at a point $X \in \mathbf{X}$, which gives a quantitative quality measure of considered ML procedure and may considered as an 'universal quality criterion'[50 - 51] in the DR.

Problems in which a reconstruction mapping is required arise in numerous applications.

**Example 1** concerns a *Wing shape optimization* problem, which is one of important problems in aircraft designing. Design variables include a number of high-dimensional vectors X of dimensions p which are detailed descriptions of wing airfoils. In practical applications, the dimension p varies in the range from 50 to 200; specific value of p is selected depending on a required accuracy of airfoil description.

Low-dimensional airfoil parameterization[74] is usually constructed in order to reduce the number of design variables, and the ME technique is one of highly powerful methods for such parameterization[9], [11]. The ME technique constructs an Embedding mapping h(3) (low-dimensional parameterization) based on given dataset consisting of high-dimensional descriptions of airfoils-prototypes and allows to describe the arbitrary airfoils from aircraft wings by low-dimensional vectors whose dimension q varies in the range from 5 to 10.

The constructed airfoil parameterization determines a low-dimensional design space, and the wing shape optimization problem is reduced to optimizing some functional on this space. If y* is the result of chosen optimization procedure in the constructed low-dimensional design space, then it is required to reconstruct a detailed description X* = g(y*) (5), (6) of the 'optimal' wing airfoil with low-dimensional descriptiom y*.   ☐

**Example 2**, described in Ref. [21], is devoted to *Processing of an Electricity price curve*. Electricity 'daily' prices are described by a multidimensional time series (electricity price curve) $X_t = (X_{t1}, X_{t2}, \ldots, X_{t,24})^T \in R^{24}$, t = 1, 2, …, T, consisting of 'hour-prices' in the course of day t. Based on $X_1, X_2, \ldots, X_T$, it is necessary to construct a forecast X* for $X_{T+1}$. Both Embedding and Reconstruction mappings are used in the proposed Forecasting algorithm[21]:

- Let the dataset $\{y_t = (y_{t1}, y_{t2}, \ldots, y_{tq})^T$, t = 1, 2, …, T} (q = 4) be the LLE-solution of the Embedding problem for the 'hour-prices' vectors $\{X_1, X_2, \ldots, X_T\}$.
- Based on a one-dimensional time series $\{y_{tk}$, t = 1, 2, …, T}, by using standard forecasting technique, a forecast $Y_k$ for $y_{T+1,k}$ is constructed, k = 1, 2, …, q.
- Based on the constructed q-dimensional vector $Y^* = (Y_1, Y_2, \ldots, Y_q)^T$, a forecast X* = g(Y*) for $X_{T+1}$ is constructed with using LLE reconstruction function g[65].

The citations[21]: 'the reconstruction of high-dimensional forecasted price curves from low-dimensional prediction is a significant step for forecasting' and 'reconstruction accuracy is critical for the application of manifold learning in the prediction'.  ☐

There are some (though limited number of) methods for a reconstruction the Data space **X** from the embedded space h(**X**). For specific linear manifold, the reconstruction can be easily made with the PCA. For nonlinear manifold, the dataset-based Replicative Neural Network[2], [35-36], [45], [57], [63] (called Auto-Encoder, AE) determines both an encoder (embedding mapping) and a decoder (reconstruction mapping), and these mappings belong to the chosen specified parametric class of the activation functions (the logistic sigmoids or hyperbolic tangents, for example). There are improved versions of the AE – the various regularized AE which are constructed under some penalties in an optimization of the activation functions parameters (the squared Frobenius norm of the encoder's Jacobian in Contractive AE[60]; a local variation of this Jacobian in Ref. [61], etc.). LLE reconstruction algorithm[65] which is done in the same manner as LLE, is another example of reconstruction mapping. LTSA reconstruction, an interpolation-like reconstruction, and nonparametric regression reconstruction, have been proposed in Ref. [85].

In this paper we propose an amplification of the ML, called the **Tangent Bundle Manifold Learning** (TBML) in which the proximity not only between the original manifold and its sample-based estimator but also between their tangent spaces is required. We present a new geometrically motivated algorithm which was first introduced in Ref. [13] that solves the TBML and gives a new solution for the ML also.

In Section 2, we propose a strict formal definition of the ML consisting in reconstruction of the D-manifold **X** (4) from the sample $\mathbf{X}_n$ which means constructing a dataset-based **Estimated Data manifold** (ED-manifold) approximating the D-manifold.

The reconstruction errors (7) can be directly computed for the sample points $X \in \mathbf{X}_n$, and for OoS points $X \in \mathbf{X} \setminus \mathbf{X}_n$ it describes the **generalization ability** of the considered ML procedure (h, g) at the point X. In Section 3, a local lower bound is obtained for the maximum reconstruction error in a small neighborhood of an arbitrary point $X \in \mathbf{X}$. The lower bound is defined in terms of the distance between tangent spaces to the D-manifold **X** and the ED-manifold at the considered point and the reconstructed point, respectively. It follows from the bound that the greater the distances between these tangent spaces, the lower the local generalization ability of the ML procedure.

Thus, it is natural to require that the ML procedure (h, g) ensures not only a proximity between the points $X \in \mathbf{X}$ and their reconstructed values X* (6) but also a proximity between the corresponding tangent spaces. A statement of such extended ML problem which includes a requirement of tangent spaces proximity and was called above as the TBML, is proposed in Section 4.

The solution of the TBML based on the proposed **Grassmann&Stiefel Eigenmaps** (GSE) approach is described in Section 5; mathematical justification for the proposed approach will be considered in subsequent version of this paper. Results of performed comparative numerical experiments with the GSE-algorithm are presented in Section 6.

## 2  Manifold Learning problem

In this paper, we focus on the following formal statement of the ML.

**ML definition.** Based on the dataset $\mathbf{X}_n$ (1) sampled from a q-dimensional D-manifold $\mathbf{X}$ in $R^p$ covered by a single chart (4), construct an **ML-solution** $\theta = (h, g)$ consisting of:

- an Embedding mapping h (3), defined on the domain $\mathbf{X}_h \supseteq \mathbf{X}$, and
- a Reconstruction mapping g (5) defined on the domain $\mathbf{Y}_g \supseteq h(\mathbf{X}_h) \supseteq h(\mathbf{X})$,

which ensures the approximate equality

$$X \approx g(h(X)) \text{ for all } X \in \mathbf{X}. \tag{8}$$

**Note.** The ML includes a construction of the domains $\mathbf{X}_h$ and $\mathbf{Y}_g$ also.

We assume that $\mathbf{X}$ is a well-behaved manifold: the mapping f is a diffeomorphism from **B** to **X** with a differentiable inverse map whose Jacobian (p×q matrix) $J_f(b)$ has Rank q for all b ∈ **B**. It is also assumed that there are positive constants C, C′, C″ such that if two arbitrary points X = f(b) and X′ = f(b′) from **X** satisfy the condition $\|X - X'\| < C$, then the inequalities

$$C' \times \|b - b'\| \leq \|X - X'\| \leq C'' \times \|b - b'\|$$

hold true. Therefore, the manifold **X** has a tube $Tube_\varepsilon(X)$ of positive radius ε (i.e., an ε-neighborhood of **X** whose points have a single projection onto **X**) whence comes that **X** has no self-intersections.

In $R^p$, the solution $\theta = (h, g)$ determines a q-dimensional manifold

$$\mathbf{X}_g = \{X = g(y) \in R^p: y \in \mathbf{Y}_g \subset R^q\} \subset R^p,$$

embedded in p-dimensional space, whose constriction

$$\mathbf{X}_\theta = \{X = g(y) \in R^p: y \in \mathbf{Y}_\theta \subset R^q\} \subseteq \mathbf{X}_g \tag{9}$$

on the **Estimated Coordinate space**

$$\mathbf{Y}_\theta = h(\mathbf{X}) \tag{10}$$

was called above as the ED-manifold. Note that

$$r_\theta = g(h(X)) \tag{11}$$

is a mapping from the D-manifold **X** to the ED-manifold, and $\mathbf{X}_\theta = r_\theta(\mathbf{X})$.

The approximate equalities (8) can be written as a **Manifold proximity** property

$$\mathbf{X} \approx \mathbf{X}_\theta, \tag{12}$$

meaning that the ED-manifold $\mathbf{X}_\theta$ accurately reconstructs the D-manifold **X**.

**Note.** As was mentioned above, the defined ML consists in constructing the **parameterized** ED-manifold (9) which approximates the initial D-Manifold (4). This problem differs from the *manifold approximation problem* consisting in a representation of the manifold geometry (manifold reconstruction) from the randomly sampled dataset by some geometrical structure in the original ambient space $R^p$, without any 'global parameterization'[42]. For the latter problem, some solutions are known, such as approximations by a simplicial complex[28] or by a finite number of affine subspaces called 'flats'[42].

From the Statistical point of view, the defined ML may be considered as a Statistical Estimation Problem: there is an unknown object **X** (4) (smooth q-dimensional parameterized D-Manifold in $R^p$ covered by a single chart) and a finite dataset $\mathbf{X}_n$ randomly sampled from **X**. Based on the sample, it is required to construct an estimator $\mathbf{X}_\theta$ (9) (also a q-dimensional parameterized ED-manifold in $R^p$ covered by a single chart) for **X** which is determined by a pair $\theta$ = (h, g) of mappings (3), (5). Quality of a solution $\theta$ = (h, g) is defined as an accuracy in the approximated equality (12), and the quantity $\delta(X)$ (7) is a quality measure at a specific point $X \in \mathbf{X}$. Some statistical aspects of the ML are considered in Ref. [38].

## 3 Local generalizing ability in Manifold Learning

As was mentioned above, the Reconstruction error $\delta_\theta(X) = \|X - r_\theta(X)\|$ can be directly computed for the sample points $X \in \mathbf{X}_n$, and for an OoS point $X \in \mathbf{X} \setminus \mathbf{X}_n$ it describes the generalization ability of the solution $\theta$ = (h, g) at the point X.

Let $X_0 \in \mathbf{X}$ be arbitrary chosen point, and let
$$\delta_\theta(X_0, \varepsilon) = \max\{\delta_\theta(X): X \in U_\varepsilon(X_0)\} \quad (13)$$
be the maximum reconstruction error in the $\varepsilon$-neighborhood
$$U_\varepsilon(X_0) = \{X \in \mathbf{X}: \|X - X_0\| \leq \varepsilon\}$$
of the point $X_0$. The quantity $\delta_\theta(X_0, \varepsilon)$ (13) characterizes the local generalization ability of the procedure $\theta$ in the neighborhood of the point $X_0$.

If the D-manifold **X** lies in a tube $\text{Tube}(\mathbf{X}_\theta)$ of the ED-manifold $\mathbf{X}_\theta$ (9), one can consider a new solution $\theta(g) = (h_g, g)$, where
$$h_g(X) = \arg\min_{y \in \mathbf{Y}}\|X - g(y)\| \text{ for } X \in \text{Tube}(\mathbf{X}_\theta) \quad (14)$$
is the projection function onto the ED-manifold $\mathbf{X}_\theta$. By definition, we have the inequality
$$\|X - r_{\theta(g)}(X)\| \leq \|X - r_\theta(X)\| \text{ for } X \in \mathbf{X}, \quad (15)$$
where $r_{\theta(g)}(X) = g(h_g(X))$.

To formulate an obtained result about a lower bound for the averaged reconstruction error (13), introduce some notations. Consider affine tangent subspaces $T(X)$ and $T_\theta(X')$ to the manifolds **X** and $\mathbf{X}_\theta$ at the points $X = f(b) \in \mathbf{X}_\theta$ and $X' = g(y') \in \mathbf{X}_\theta$, respectively. These affine subspaces of dimension q in $R^p$ can be represented in the form of direct sums
$$T(X) = X \oplus L(X),$$
$$T_\theta(X') = X' \oplus L_\theta(X'), \quad (16)$$
where
$$L(X) = \text{Span}(J_f(b)),$$
$$L_\theta(X') = \text{Span}(J_g(y')) \quad (17)$$
are linear subspaces of dimension q in $R^p$ that are spanned by columns of Jacobians $J_f(b)$ and $J_g(y')$ of the mappings f (4) and g (5), respectively. In what follows, the linear spaces L and $L_\theta$ will be treated as elements of the Grassmann manifold[84] Grass(p, q) composed of all the q-dimensional linear subspaces in $R^p$.

For the elements L, L' $\in$ Grass(p, q), the quantity
$$d_{P,2}(L, L') = \|P_L - P_{L'}\|_2 = \sin\zeta_{\max}(L, L') \quad (18)$$
is a metric (called the projection metric in 2-norm[80] or simply the projection 2-norm[27], [33]) on the Grassmann manifold; here $P_L$ and $P_{L'}$ are projectors onto the linear spaces L and L', and $\zeta_{\max}(L, L')$ is the maximum principal angle[30], [37], [39] between the subspaces L and L'. In Statistics, metric (18) is called the Min Correlation Metric[33].

The following theorem and its corollary hold true.

**Theorem 1.** If h and g are smooth full-rank mappings and if the D-manifold **X** lies in the tube Tube($\mathbf{X}_\theta$) of the Empirical manifold $\mathbf{X}_\theta$, then the following inequality for the local maximum reconstruction error $\delta_{\theta(g)}(X_0, \varepsilon)$ holds as $\varepsilon \to 0$:

$$\delta^2_{\theta(g)}(X_0,\varepsilon) \geq \delta^2_{\theta(g)}(X_0) + \varepsilon^2 \times d^2_{P,2}(L(X_0), L_{\theta(g)}(r_{\theta(g)}(X_0)) + o(\varepsilon^2); \quad (19)$$

where $L_{\theta(g)}(r_{\theta(g)}(X)) = \text{Span}(J_g(h_g(X)))$; hereafter, the o($\cdot$) symbol in the vector case is understood componentwise. For $\delta_{\theta(g)}(X_0) = 0$, inequality (19) turns into equality.

**Note.** Taking into account (15), Theorem 1 establishes a lower bound for the quantity $\delta_\theta(X_0, \varepsilon)$ (13).

*Proof:* Let $X \in U_\varepsilon(X_0)$, and $X = f(b)$, $X_0 = f(b_0)$, where $b, b_0 \in \mathbf{B}$. Denote

$$y = h_g(X), y_0 = h_g(X_0).$$

Then the Taylor formula yields

$$r_{\theta(g)}(X) = r_{\theta(g)}(X_0) + J_g(y_0) \times (y - y_0) + o(X - X_0),$$

and it follows from (14) that

$$y = y_0 + ((J_g(y_0))^T \times J_g(y_0))^{-1} \times (J_g(y_0))^T \times (X - r_{\theta(g)}(X_0)) + o(X - X_0).$$

Let

$$J_g(y) = Q_g(y) \times \text{Diag}_g(y) \times (V_g(y))^T$$

be a Singular Value Decomposition (SVD) of the p×q matrix $J_g(y)$, where $Q_g(y)$ is a p×q orthogonal matrix. Then

$$r_{\theta(g)}(X) = r_{\theta(g)}(X_0) + \pi(y_0) \times (X - r_{\theta(g)}(X_0)) + o(X - X_0), \quad (20)$$

where

$$\pi(y_0) = Q_g(y_0) \times (Q_g(y_0))^T$$

is the projector onto the linear space $L_{\theta(g)}(r_{\theta(g)}(X_0))$ (17).

It also follows from (14) that

$$(X_0 - r_{\theta(g)}(X_0)) \in (L_{\theta(g)}(r_{\theta(g)}(X_0)))^\perp;$$

hence,

$$\pi(y_0) \times (X - r_{\theta(g)}(X_0)) = \pi(y_0) \times (X - X_0),$$

and (20) takes the form

$$r_{\theta(g)}(X) = r_{\theta(g)}(X_0) + \pi(y_0) \times (X - X_0) + o(X - X_0),$$

whence comes the relation

$$X - r_{\theta(g)}(X) = (X_0 - r_{\theta(g)}(X_0)) + \pi^\perp(y_0) \times (X - X_0) + o(X - X_0), \quad (21)$$

where $\pi^\perp(y_0)$ is the projector onto the linear space $(L_{\theta(g)}(r_{\theta(g)}(X_0)))^\perp$. Let

$$J_f(b) = Q_f(b) \times \text{Diag}_f(b) \times (V_f(b))^T$$

be the SVD-decomposition of the p×q matrix $J_f(b)$, where $Q_f(b)$ is a p×q orthogonal matrix.

Consider a q×q matrix $(Q_g(y_0))^T \times Q_f(b_0)$ such that its SVD-decomposition has the form

$$(Q_g(y_0))^T \times Q_f(b_0) = O_1 \times \text{Diag}(\cos(\zeta)) \times (O_2)^T,$$

where $O_1$ and $O_2$ are q×q orthogonal matrices and where the diagonal elements of the diagonal matrix

$$\text{Diag}(\cos(\zeta)) = \text{Diag}(\cos(\zeta_q), \cos(\zeta_{q-1}), \ldots, \cos(\zeta_1))$$

are cosines of the principal angles[30] between the subspaces $L(X_0)$ and $L_{\theta(g)}(r_{\theta(g)}(X_0))$ arranged in ascending order:

$$0 \leq \zeta_1 \leq \zeta_2 \leq \ldots \leq \zeta_q \leq \pi/2.$$

The columns $\{t_{f,1}, t_{f,2}, \ldots, t_{f,q}\}$ and $\{t_{g,1}, t_{g,2}, \ldots, t_{g,q}\}$ of the p×q orthogonal matrices $Q_f(b_0) \times O_2$ and $Q_g(y_0) \times O_1$ are principal vectors in the subspaces $L(X_0)$ and $L_{\theta(g)}(r_{\theta(g)}(X_0))$. These vectors determine the orthonormal bases for these subspaces and satisfy the relations

$$(t_{f,i}, t_{g,j}) = \delta_{ij} \times \cos(\zeta_{q+1-j}), i, j = 1, 2, \ldots, q.$$

Taking into account the Taylor series expansion

$$X = X_0 + J_f(b_0) \times (b - b_0) + o(b - b_0), \quad (22)$$

and the assumptions about the chart f, we obtain

$$\pi^\perp(X - X_0) = \sum_{j=1}^q t_{f,g,j} \times \alpha_j(X) + o(X - X_0),$$

where

$$t_{f,g,j} = \pi^\perp(y_0) \times t_{f,j} = t_{f,j} - t_{g,j} \times \cos\zeta_{q+1-j}, j = 1, 2, \ldots, q,$$

are projections of the principal vectors $\{t_{f,1}, t_{f,2}, \ldots, t_{f,q}\}$ onto the subspace $(L_{\theta(g)}(r_{\theta(g)}(X_0)))^\perp$, and

$$\alpha(X) = (O_2)^T \times D_f(b_0) \times (V_f(b_0))^T \times (b - b_0) \equiv (\alpha_1(X), \alpha_2(X), \ldots \alpha_q(X))^T.$$

Then it follows from (22) that the vector $\alpha(X)$ satisfies the relation

$$\max\{\|\alpha(X)\|, X \in U_\varepsilon(X_0)\} = \varepsilon + o(\varepsilon).$$

Taking into account the introduced notation and obtained relations, we get from (21) that

$$\delta^2_{\theta(g)}(X) = \sum_{j=1}^{q} \left\{A_j(X_0) + \alpha_j(X) \times \sin(\zeta_{q+1-j})\right\}^2 + o(\|X-X_0\|^2),$$

where $A_1(X_0), A_2(X_0), \ldots A_q(X_0)$ are the components of the vector $\pi^\perp(y_0) \times (X_0 - r_{\theta(g)}(X_0))$.

It can be shown that the relation

$$\max\left\{\sum_{j=1}^{q}\left\{A_j(X_0) + \alpha_j(X) \times \sin(\zeta_{q+1-j})\right\}^2, X \in U_\varepsilon(X_0)\right\} + o(\|X-X_0\|^2) =$$
$$= \max\left\{\sum_{j=1}^{q}\left\{|A_j(X_0)| + |\alpha_j(X)| \times \sin(\zeta_{q+1-j})\right\}^2, X \in U_\varepsilon(X_0)\right\} + o(\varepsilon^2) \geq$$
$$\geq \delta^2_{\theta(g)}(X_0) + \max\left\{\sum_{j=1}^{q} \alpha_j^2(X) \times \sin^2(\zeta_{q+1-j}), |\alpha(X)| \leq \varepsilon\right\} + o(\varepsilon^2),$$

is valid, whence it follows that

$$\delta^2_{\theta(g)}(X_0, \varepsilon) \geq \delta^2_{\theta(g)}(X_0) + \varepsilon^2 \times \sin^2\zeta_{max}(L(X_0), L_{\theta(g)}(r_{\theta(g)}(X_0))) + o(\varepsilon^2),$$

which proves Theorem 1. □

From the proof of Theorem 1 given above, one can derive the following corollary.

**Corollary of Theorem 1.** For $X \in U_\varepsilon(X_0)$ and $\varepsilon \to 0$, we have the following asymptotic inequalities:

$$\|(X-X_0) - (r_{\theta(g)}(X)-r_{\theta(g)}(X_0))\| \leq \|X-X_0\| \times d_{P,2}(L(X_0), L_{\theta(g)}(r_\theta(X_0))) + o(\|X-X_0\|) \quad (23)$$

and

$$\|X-X_0\| \times \sqrt{1 - d^2_{P,2}(L(X_0), L_{\theta(g)}(r_{\theta(g)}(X_0)))} + o(\|X-X_0\|) \leq$$
$$\leq \|r_{\theta(g)}(X) - r_{\theta(g)}(X_0)\| \leq \|X-X_0\| + o(\|X-X_0\|). \quad (24)$$

The inequality (23) indicates to what extent the mapping $r_{\theta(g)}$ preserves the local structure of the Data manifold, while the inequality (24) characterizes the local non-isometricity of this mapping.

## 4 Tangent Bundle Manifold Learning

It follows from the above formulas that the greater the distances between the linear spaces $L(X)$ and $L_\theta(r_\theta(X))$ at the points $X$, the lower the local generalization ability of the solution $\theta$ becomes, the poorer the local structure of the D-manifold is preserved, and the poorer the local isometricity properties are ensured.

Thus, it is natural to require that the procedure $\theta$ ensures not only proximity (8) between the points $X \in \mathbf{X}$ and their images $r_\theta(X) \in \mathbf{X}_\theta$ but also proximity

$$L(X) \approx L_\theta(r_\theta(X)) \quad (25)$$

between the tangent spaces $L(X), L_\theta(r_\theta(X)) \in Grass(p, q)$ for all $X \in \mathbf{X}$ in the selected metric on the Grassmann manifold $Grass(p, q)$. The approximate equalities (25) may be treated as the **Tangent proximity** between manifolds $\mathbf{X}_\theta$ and $\mathbf{X}$.

In the Manifold theory[49], [52], the set

$$TB(\mathbf{X}) = \{(X, L(X)): X \in \mathbf{X}\} \quad (26)$$

composed of the manifold points equipped by the tangent spaces at these points is known as the **tangent bundle** of the manifold $\mathbf{X}$.

In the introduced terms, the manifold proximity (12) and the tangent proximity (25) can be referred to as the **Tangent Bundle Proximity**

$$TB(\mathbf{X}) \approx TB(\mathbf{X}_\theta), \quad (27)$$

where

$$TB(\mathbf{X}_\theta) = \{(X', L_\theta(X')): X' \in \mathbf{X}_\theta\} \equiv \{(r_\theta(X), L_\theta(r_\theta(X))): X \in \mathbf{X}\} \quad (28)$$

is the tangent bundle of ED-manifold $\mathbf{X}_\theta$ (9). So, the ML solution θ must ensure the Tangent Bundle Proximity (27).

The requirement of Tangent Bundle Proximity in the ML arises in various applications where a ML-solution is an intermediate step in considered Data Analysis problem.

**Example 3.** Consider a problem consisting in optimization of some functional υ(X) depending on a p-dimensional vector X lying on a q-dimensional manifold **X** (4) in $R^p$, q < p, covered by a single chart f. By definition, this problem is equivalent to the problem of optimization of a functional v(b) ≡ υ(f(b)) defined on a q-dimensional domain (Coordinate space) **B**.

In applications, an analytical description of Manifold **X** may be unknown, and only a sample $\mathbf{X}_n$ from **X** is available (see Example 1 concerning Wing shape optimization). Based on some solution θ = (h, g) for the ML, the manifold **X** is approximated by an ED-manifold $\mathbf{X}_\theta$ (9) providing the approximate equalities (8), (12), whence comes

$$\upsilon(r_\theta(X)) \approx \upsilon(X). \quad (29)$$

The latter relation (29) is ensured if θ is a solution of the **Functional DR** (FDR) Problem stated in Ref. [47], in which the ML-solution θ must provide not only the proximity (8) but also the **Functional proximity** $F(X) \approx F(r_\theta(X))$ for a specified family of functions {F}; a neural network-based solution for the FDR was obtained in Ref. [10].

Thus, an initial optimization problem can be replaced by optimization of the functional υ on $\mathbf{X}_\theta$, or, equivalently, by optimization of the empirical functional $\upsilon_\theta(X) = \upsilon(r_\theta(X))$. But

$$\upsilon(r_\theta(X)) = \upsilon(g(h(X))) = \upsilon_\theta(y) \equiv \upsilon(g(y)), \; y = h(X),$$

and the initial problem reduces to optimization in a q-dimensional Empirical coordinate space $\mathbf{Y}_\theta = h(\mathbf{X}) \subset R^q$.

To ensure a closeness between the iterative optimization processes induced by the same optimization gradient-based method applied to the functionals v(b) and $v_\theta(y)$, respectively, it is required to guarantee both an accurate reconstruction of an initial design space **X** (12), (29) and an accurate reconstruction of its tangent spaces (25). □

The problem of accurate reconstruction of the tangent bundle TB(**X**) from the sample $\mathbf{X}_n$ arises in various applications.

**Example 4.** Many tracking problems in control and computer vision deal with the observed trajectories lying on a low-dimensional manifold embedded in a high-dimensional space, and a Learning manifold shape is a necessary step for the ML, which, in turn, allows faster and robust tracking performances[69]. Any dynamic system defined on the D-manifold **X** must induce the trajectories whose both the velocities and their points of applivations belong to TB(**X**)[58]. Various motion tracking problems from video sequences require[69 – 71] a constructing an approximation for the tangent bundle TB(**X**) of the D-manifold **X** (4). The latter problem consists in dataset-based constructing the tangent bundle of ED-manifold which approximates the TB(**X**).

In this paper, we focus on the following formal statement of the TBML.

**Tangent Bundle Manifold Learning definition.** Based on a dataset $\mathbf{X}_n$ (1) sampled from a q-dimensional D-Manifold **X** in $R^p$ (4) with tangent bundle TB(**X**) (26), construct a triple (TBML-solution) ϑ = (h, g, G) = (θ, G) consisting of:

- a solution θ = (h, g) for the ML which ensures the manifold proximity (12);
- a p×q matrix G(y) with Rank G(y) = q defined on a domain $\mathbf{Y}_g \supset \mathbf{Y}_\theta = h(\mathbf{X})$,

to ensure both: the equalities:

$$G(y) = J_g(y) \text{ for all } y \in \mathbf{Y}_\theta, \quad (30)$$

and the proximities:

$$L(X) \approx \text{Span}(G(h(X))) \text{ for all } X \in \mathbf{X}. \quad (31)$$

It follows from (30) that

$$\text{Span}(G(h(X))) = \text{Span}(J_g(h(X))) = L_\theta(r_\theta(X)) \text{ for all } X \in \mathbf{X}, \tag{32}$$

and

$$TB(\mathbf{X}_\theta) = \{(g(y), \text{Span}(G(y))), y \in \mathbf{Y}_\theta\}, \tag{33}$$

whence comes the tangent bundle proximity (27). Denote

$$\mathbf{L} = \{L(X), X \in \mathbf{X}\} \subset \text{Grass}(p, q), \tag{34}$$

the q-dimensional submanifold in the Grassmann manifold Grass(p, q) called **Tangent Manifold**. The TBML-solution $\vartheta = (h, g, G)$ determines a dataset-based q-dimensional submanifold

$$\mathbf{L}_{g,G} = \{\text{Span}(G(y)), y \in \mathbf{Y}_g \subset R^q\} \subset \text{Grass}(p, q)$$

whose constriction

$$\mathbf{L}_\vartheta = \{\text{Span}(G(y)), y \in \mathbf{Y}_\theta \subset R^q\} \subset \text{Grass}(p, q)$$

on the Estimated Coordinate space $\mathbf{Y}_\theta = h(\mathbf{X})$ coincides with submanifold

$$\mathbf{L}_\theta = \{L_\theta(X'): X' \in \mathbf{X}_\theta\} \equiv \{L_\theta(r_\theta(X)): X \in \mathbf{X}\} \subset \text{Grass}(p, q) \tag{35}$$

and approximates the Tangent Manifold (34):

$$\mathbf{L}_\vartheta = \mathbf{L}_\theta \approx \mathbf{L}. \tag{36}$$

The dataset-based submanifold $\mathbf{L}_\vartheta = \mathbf{L}_\theta$ (35) can be referred to as the **Estimated Tangent Manifold**.

From the Statistical point of view, the defined TBML may also be considered as a Statistical Estimation Problem: from a finite dataset $\mathbf{X}_n$ randomly sampled from a smooth q-dimensional parameterized D-manifold $\mathbf{X}$ in $R^p$, to estimate its tangent bundle TB($\mathbf{X}$) (26).

**Note.** The notion 'tangent bundle' is used in the ML for various purposes: as an approximation of a manifold shape from the data in Ref. [28]; for geometric interpretation of the Contractive Auto-Encoder[60] in Ref. [62]; in discussing about the geomerric aspects of the ML and Subspace Leraning[16] methods in Ref. [43]; in a name of the 'Tangent Bundle Approximation' algorithm for the Tangent Space Learning Problem[34] in Ref. [69], [71 – 72], etc. The above given TBML definion is new and it differs from all the known usages of this term in ML.

## 5 Grassman/Stiefel Eigenmaps algorithm

The proposed solution for the TBML called Grassman&Stiefel Eigenmaps consists of several stages. After a standard auxiliary stage (Stage 1), a family **H** consisting of the p×q matrices H(X) smoothly depending on $X \in R^p$ is constructed, such that the submanifold $\mathbf{L}_H = \{L_H(X) = \text{Span}(H(X)), X \in \mathbf{X}\} \subset \text{Grass}(p, q)$ approximates the Tangent manifold **L** (34) (Stage 2). After that, we construct the Embedding mapping h (3) (Stage 3) and the Reconstruction mapping g (5) (Stage 4) which in total determine the ED-manifold $\mathbf{X}_\theta$. These mapping are constructed such a way that the ED-manifold $\mathbf{X}_\theta$ approximates the D-manifold $\mathbf{X}$ and whose Estimated Tangent Manifold $\mathbf{L}_\theta$ (35) coincides with the submanifold $\mathbf{L}_H$, whence comes the Tangent Bindle Proximity (27).

Sections 5.1 – 5.4 describe the Stages 1 – 4, respectively. Section 5.5 contain some properties of the proposed GSE algorithm, Section 5.6 describes an 'orthogonal' version of the GSE algorithm.

Note that the presented GSE-algorithm will not attempt to deal with two problems which plague any manifold learning algorithm: a noise and an undersampling. From a practical point of view, a manifold learning algorithm must be able to address both problems when dealing with real data sets. However, given arbitrary and unknown geometry and topology, the considered problem of the TBML from noiseless and sufficiently dense data is a still very difficult challenge. The proposed algorithm may be thought of as a first step, and the ultimate goal would be to extend the framework to more realistic and challenging cases which involve these two problems.

### 5.1 Nearness measures and linear PCA spaces.

The following steps contain the details of Stage 1. By numbers $\{\varepsilon_i > 0, i = 1, 2, \ldots\}$ we denote algorithm parameters.

**Step1.1 [Euclidean nearness].** For $X \in \mathbf{X}_n$, put $U_E(X) = \{X' \in \mathbf{X}_n: \|X' - X\| < \varepsilon_1\}$; for an OoS point $X \notin \mathbf{X}_n$, put $U_E(X) = \{X' \in \mathbf{X}_n: \|X' - X\| < \varepsilon_1\} \cup \{X\}$. Following Ref. [3], define the Euclidean kernel

$$K_E(X, X') = K_0(X, X') \times \exp\{-\varepsilon_2 \times \|X - X'\|^2\}, \tag{37}$$

where $K_0(X, X') = 1$ if $X \in U_E(X')$ and $X' \in U_E(X)$, and $K_E(X, X') = 0$ otherwise.

**Step1.2 [PCA-based tangent spaces].** By applying the Weighted PCA[46] with weights $K_E$ (37) to the set $U_E(X)$, the ordered eigenvalues $\lambda_1(X) \geq \lambda_2(X) \geq \ldots \geq \lambda_p(X)$ and the corresponding orthonormal principal vectors are constructed. Denote the set

$$\mathbf{X}_h = \{X \in \mathbb{R}^p: \lambda_q(X) > \varepsilon_3\},$$

which will be a domain of the subsequently constructed Embedding mapping h.

For $X \in \mathbf{X}_h$, define a $p \times q$ matrix $Q_{PCA}(X) \in \text{Stief}(p, q)$ with the columns consisting of the first q principal vectors; here Stief(p, q), referred to as the Stiefel manifold[27], [52] consists of all the orthogonal $p \times q$ matrices. Denote by

$$L_{PCA}(X) = \text{Span}(Q_{PCA}(X)) \in \text{Grass}(p, q) \tag{38}$$

the linear space spanned by columns of $Q_{PCA}(X)$.

In what follows, we assume that the D-manifold **X** is well sampled to provide an inclusion $\mathbf{X}_n \subset \mathbf{X}_h$. If $X \in \mathbf{X}_h \cap \mathbf{X}$ and the neighborhood $U_E(X)$ is small enough, then[1]

$$L(X) \approx L_{PCA}(X). \tag{39}$$

**Step1.3 [Grassmannian nearness].** Denote

$$S(X, X') = (Q_{PCA}(X))^T \times Q_{PCA}(X'), \tag{40}$$

and introduce the Binet-Cauchy metric[33], [83]

$$d_{BC}(L_{PCA}(X), L_{PCA}(X')) = (1 - \text{Det}^2(S(X, X')))^{1/2} \tag{41}$$

on the Grassmann manifold which induces the Binet-Cauchy kernel[33]

$$K_{BC}(L_{PCA}(X), L_{PCA}(X')) = \text{Det}^2(S(X, X')).$$

**Note.** Among all the known metrics on a Grassmann manifold, there are only two metrics that induce corresponding kernels on the Grassmann manifold: the Binet-Cauchy metric (41) and the Projection metric[33] $d_{P,F}(L, L') = 2^{-1/2} \times \|P_L - P_{L'}\|_F$, also called the Projection F-norm[27] and the Chordal distance[22] between L and L′. There are some reasons to use namely the Binet-Cauchy metric in the GSE-algorithm.

**Note.** It is possible to use an iterative version for Step 1.2 in which the neighborhood $U_E(X)$ is used in the starting iteration only and is replaced by the neighborhood

$$U_G(X) = \{X' \in U_E(X): d_{BC}(L_{emp}(X), L_{emp}(X')) < \varepsilon_4\}$$

in the next iterations; the PCA is applied to the points from the current neighborhood.

**Note.** In Ref. [34], [69 − 72], [79] and others, various methods for a choice of a neighborhood for applying the PCA are proposed. In some of these papers, the neighborhoods like $U_G(X)$ based on the principal angles are used.

**Step1.4 [Aggregate kernel].** Define the aggregate kernel

$$K(X, X') = K_E(X, X') \times K_G(X, X') \tag{42}$$

for points $X \in \mathbf{X}_h$ and $X' \in \mathbf{X}_n$. This kernel reflects not only a nearness between the points X and X′ but also a nearness between the linear spaces $L_{PCA}(X)$ and $L_{PCA}(X')$, whence (39) comes a nearness between the tangent spaces $L(X)$ and $L(X')$.

## 5.2 *Estimated Tangent Manifold.*

Denote Stief-NC(p, q) the non-compact Stiefel manifold[27], [52] consisting of all the tall-skinny p×q matrices M, q ≤ p, with Rank(M) = q. In Stage 2, a smooth mapping

$$H: X \in \mathbf{X} \to H(X) \in \text{Stief-NC}(p, q) \tag{43}$$

from the D-manifold **X** to the non-compact Stiefel manifold Stief-NC(p, q) is constructed such that ensures a nearness

$$L_H(X) \approx L(X) \text{ for all } X \in \mathbf{X} \tag{44}$$

between the linear spaces $L_H(X) = \text{Span}(H(X)) \in \text{Grass}(p, q)$ and $L(X)$, and optimizes also some below defined cost function.

To achieve (44), we will construct the matrices $H(X)$, $X \in \mathbf{X}_h$, such to satisfying the relation

$$L_H(X) = L_{PCA}(X); \tag{45}$$

then the desired relation (44) follows from (39). The Grassmann manifold Grass(p, q) can be considered as a quotient space of the compact Stiefel manifold Stief(p, q) by the group Stief-NC(q, q)[27], so matrices $H(X)$ can be presented in the form

$$H(X) = Q_{PCA}(X) \times v(X), \tag{46}$$

with $v(X) \in \text{Stief-NC}(q, q)$.

A constructing the matrices $H(X)$ (46) consist of two steps. First, a preliminary matrix set

$$\mathbf{H}_n = \{H_i = Q_{PCA}(X_i) \times v_i, X_i \in \mathbf{X}_n\}, \tag{47}$$

where $v_1, v_2, \ldots, v_n$ are the q×q matrices, is constructed to minimize the quadratic form

$$\Delta_H(\mathbf{H}_n) = \tfrac{1}{2} \sum_{i,j=1}^n K(X_i, X_j) \times \|H_i - H_j\|_F^2 \tag{48}$$

under the normalizing conditions

$$\sum_{i=1}^n K_i \times (H_i^T \times H_i) = I_q = I_q, \tag{49}$$

where $I_q$ is the q×q identity matrix, and

$$K_i = \tfrac{1}{K} \sum_{j=1}^n K(X_i, X_j), i = 1, 2, \ldots, n; K = \sum_{i,j=1}^n K(X_i, X_j). \tag{50}$$

From (46), the quadratic form (48) can be written as

$$\Delta_V(\mathbf{V}_n) = \tfrac{1}{2} \sum_{i,j=1}^n K(X_i, X_j) \times \|Q_{PCA}(X_i) \times v_i - Q_{PCA}(X_j) \times v_j\|_F^2, \tag{51}$$

and a minimizing of each (i, j)-summand in (51) over the matrices $v_i$ and $v_j$ is known as the Procrustes problem[27],[32],[77], here the (nq)×q matrix

$$\mathbf{V}_n = \left(v_1^T : v_2^T : \cdots : v_n^T\right)^T, \tag{52}$$

is composed of the q×q matrices $v_1, v_2, \ldots, v_n$. So, a minimizing (51) over $\mathbf{V}_n$ may be referred to as the Averaged Procrustes problem.

Under the normalizing condition (49), the sum (51) can be written as

$$\Delta_V(\mathbf{V}_n) = K - \sum_{i,j=1}^n K(X_i, X_j) \times \text{Tr}(v_i^T \times S(X_i, X_j) \times v_j),$$

whence the optimization problem (48), (51) reduces to maximizing the quadratic form

$$\Delta^*(\mathbf{V}_n) = \text{Tr}(\mathbf{V}_n^T \times \Phi_1(S) \times \mathbf{V}_n). \tag{53}$$

under the normalizing condition

$$\mathbf{V}_n^T \times \Phi_0 \times \mathbf{V}_n = I_q, \tag{54}$$

where the nq×nq matrices $\Phi_1(S) = \|\Phi_{ij}(S)\|$ and $\Phi_0 = \|\Phi_{ij}\|$ consist of q×q matrices:

$$\Phi_{ij}(S) = K(X_i, X_j) \times S(X_i, X_j); \Phi_{ij} = \delta_{ij} \times K_i \times I_q,$$

here δ is Kronecker's symbol, i, j = 1, 2, …, n.

The latter optimization problem (53), (54), in turn, reduces to solving the generalized eigenvector problem

$$\Phi_1(S) \times \mathbf{V}_n = \lambda \times \Phi_0 \times \mathbf{V}_n, \tag{55}$$

whose solution defines the following Step.

**Step2.1 [Generalized eigenvector problem].** Let $V_1, V_2, \ldots, V_q \in R^{nq}$ be the orthonormal eigenvectors corresponding to the q largest eigenvalues in (55), and the (nq)×q matrix

$$\mathbf{V}_n^* = (V_1 : V_2 : \ldots : V_q) \tag{56}$$

is composed of these vectors as their columns. Define the set

$$\mathbf{H}_n^* = \{H_1^*, H_2^*, \ldots, H_n^*\} \tag{57}$$

consisting of the matrices

$$H_i^* = Q_{PCA}(X_i) \times v_i^*, \tag{58}$$

where $v_1^*, v_2^*, \ldots, v_n^*$ are q×q submatrices of $\mathbf{V}_n^*$ induced by the representation (52).

After the First step, under the constructed set $\mathbf{H}_n^*$ (57), the value H(X) (46) for an arbitrary point X is constructed by minimizing the quadratic form

$$\Delta_H(H(X)) = \tfrac{1}{2}\sum_{j=1}^n K(X, X_j) \times \|H(X) - H_j\|_F^2 \tag{59}$$

over the q×q matrix v(X). Such matrix H(X) can be obtained in an explicit form written in the following step.

**Step2.2 [Aligning the tangent spaces].** Let $X \in \mathbf{X}_h$ be an arbitrary point. Given the set $\mathbf{H}_n^*$ (57) from Step 1.1, compute the matrix

$$H(X) = H(X|\mathbf{H}_n^*) = \pi(X) \times H_{KNR}(X) \tag{60}$$

which minimizes the quadratic form (59), where

$$\pi(X) = Q_{PCA}(X) \times (Q_{PCA}(X))^T \tag{61}$$

is the projector onto $L_{PCA}(X)$ (38), and

$$H_{KNR}(X) = \sum_{j=1}^n K^*(X, X_j) \times H_j^* \tag{62}$$

is a standard Kernel Non-parametric Regression[81] estimator for H(X) based on the preliminary values $H_j^*$ (58) of the matrix H(X) in the sample points, here

$$K^*(X, X_j) = \frac{K(X, X_j)}{K(X)}, j = 1, 2, \ldots, n; K(X) = \sum_{j=1}^n K(X, X_j).$$

The solution (60) – (62) can be written in the form (46) with the matrix

$$v(X) = \sum_{j=1}^n K^*(X, X_j) \times S(X, X_j) \times v_j^*. \tag{63}$$

The solution H(X) (60) – (62) determines the family

$$\mathbf{L}_H = \{L_H(X) = \text{Span}(H(X)), X \in \mathbf{X}\} \subset \text{Grass}(p, q) \tag{64}$$

consisting of the linear spaces $L_H(X)$ smoothly depending on $X \in \mathbf{X}$. It follows from (39) and (45) that the family $\mathbf{L}_H$ (64) approximates the Tangent manifold **L** (34) and is a result of an alignment of the PCA-based linear spaces $\{L_{PCA}(X)\}$ (38).

**Note.** The problem of an estimating the tangent spaces L(X) (17) in a form of a function of point $X \in \mathbf{X}$ was considered in some previous works: the matrices whose rows span the tangent spaces were constructed by using an Artificial Neural Networks with one hidden layer[6] or a Radial Basic Functions[24 - 25]. In Ref. [34] the other Grassmann manifold-based method (PTSL) for constructing the tangent spaces which smoothly varied on the manifold is proposed. In Ref. [69 – 72] the aligned tangent spaces for all the manifold points are constructed using the proposed Tangent Bundle Approximation[69 – 71] algorithm or its generalization[72].

**Note.** In general case, the final values $\{H(X_i) = H(X_i|\mathbf{H}_n^*)\}$ (60) – (62) at the sample points do not coincide with the preliminary values $\{H_i^*\}$ (57), (58). But the set (57) computed in the Step 2.1 minimizes also the averaged residual

$$D(\mathbf{H}_n) = \sum_{i=1}^n K_i \times \|H_i - H(X_i|\mathbf{H}_n)\|_F^2$$

over the sets $\mathbf{H}_n$ (47) under the normalizing condition (49); this resudial can be written also in the form:

$$D_V(\mathbf{V}_n) = \sum_{i=1}^n K_i \times \|v_i - \sum_{j=1}^n K^*(X_i, X_j) \times S(X_i, X_j) \times v_j\|_F^2.$$

**Note.** The desired mappings h (3) and g (5) will be constructed in Stage 3 such that the Jacobian $J_g(h(X))$ of the mapping g(y) at the point y = h(X) is close to the matrix H(X), thus

$$\|H(X') - H(X)\|_F^2 \approx \sum_{k=1}^{p} |Hess_{g,k}(h(X)) \times (h(X') - h(X))|^2,$$

where $Hess_{g,k}$ is the Hessian of the k-th component $g_k$ from the vector function $g = (g_1, g_2, \ldots, g_p)^T$. These mappings determine the ED-manifold $X_\theta$ (9), and the minimum values of the quadratic forms (48), (59) characterize an averaged local curvature of the ED-manifold.

## 5.3 Embedding mapping

Under a given mapping H (60) – (62) constructed at Stage 2, we desire constructing the mappings h and g to provide both the equalities $g(h(X)) \approx X$ (8) and
$$J_g(h(X)) = H(X) \text{ for all } X \in X, \qquad (65)$$
whence comes
$$g(h(X')) - g(h(X)) \approx H(X) \times (h(X') - h(X)) \qquad (66)$$
for the near points $X, X' \in X$. Due the condition (8), the equation (66) can be written as
$$X' - X \approx H(X) \times (h(X') - h(X)), \qquad (67)$$
and these equalities written for all the pairs of near points $X, X' \in X_n$ can be considered as the regression equations for unknown quantities $h(X_i)$, $i = 1, 2, \ldots, n$.

A constructing the vectors $h(X)$ consist of two steps. First, the vector set
$$\mathbf{h}_n = \{h_1, h_2, \ldots h_n\} \qquad (68)$$
consisting of the preliminary values of the mapping $h(X)$ at the sample points $X_n$ is constructed to minimize the weighted residual
$$\Delta_h(\mathbf{h}_n) = \tfrac{1}{2}\sum_{i,j=1}^{n} K(X_i, X_j) \times |(X_j - X_i) - H(X_i) \times (h_j - h_i)|^2 =$$
$$= \tfrac{1}{2}\sum_{i,j=1}^{n} K(X_i, X_j) \times \{|\pi^\perp(X_i) \times (X_j - X_i)|^2 + |v_i \times (h_j - h_i) - c_{j|i}|^2\} \qquad (69)$$
under the natural normalizing condition
$$h_1 + h_2 + \ldots + h_n = \mathbf{0} \in R^q, \qquad (70)$$
where $c_{j|i} = (Q_{PCA}(X_i))^T \times (X_j - X_i)$ are the expansion coefficients of the projection of the vectors $(X_j - X_i)$ onto the linear space $L_{PCA}(X_i)$ (38) in the PCA basis $Q_{PCA}(X_i)$. A standard least squares solution of this regression problem defines the following step.

**Step3.1 [Solution of regression problem].** Compute the vector set $\mathbf{h}_n$ as a solution of the linear least squares equations for the regression problem (67), (69) with $j^{th}$ equation
$$\sum_{i=1}^{n} K(X_i, X_j) \times (v_i^T \times v_i + v_j^T \times v_j) \times (h_j - h_i) = \sum_{i=1}^{n} K(X_i, X_j) \times (v_i^T \times c_{j|i} - v_j^T \times c_{i|j}), \quad (71)$$
$j = 1, 2, \ldots, n$, complemented by the normalizing equation (70).

Then, under the given vector set $\mathbf{h}_n$ (68), the value $h(X)$ for arbitrary point $X \in X_h$ is constructed by minimizing the quadratic form
$$\Delta_h(h(X)) = \tfrac{1}{2}\sum_{j=1}^{n} K(X, X_j) \times |(X_j - X) - H(X) \times (h_j - h(X))|^2, \qquad (72)$$
which is the weighted residual for the approximate relations (67) written for X and all the near points $X' \in X_n$. A solution of the optimization problem (72) may be obtained in an explicit form which gives the following step.

**Step3.2 [Embedding mapping].** For an arbitrary point $X \in X_h$ under the given vector set $\mathbf{h}_n$, compute the vector
$$h(X) = \sum_{j=1}^{n} K^*(X, X_j) \times h_j + v^{-1}(X) \times (Q_{PCA}(X))^T \times (X - \sum_{j=1}^{n} K^*(X, X_j) \times X_j) \qquad (73)$$
which minimizes the residual (72), here $v(X)$ is defined in (63).

It follows from the definition (73), a Jacobian of the mapping $h(X)$ is
$$J_h(X) = v^{-1}(X) \times (Q_{PCA}(X))^T. \qquad (74)$$

**Note.** The first summand in (73)
$$h_{KNR}(X) = \sum_{j=1}^{n} K^*(X, X_j) \times h_j \qquad (75)$$
is a standard Kernel Non-parametric Regression-based estimator for $h(X)$ based on the preliminary values $h_j \in \mathbf{h}_n$ of the vector $h(X)$ at the sample points.

## *5.4 Constructing the Reconstruction mapping*

At Stage 4, a nearness measure in Estimated Coordinate space and Reconstruction mapping g together with its Jacobian $G = J_g$ are constructed in subsections 5.4.1 and 5.4.2, respectively. This mapping together with the constructed Embedding mapping (73) determine the ED-manifold $\mathbf{X}_\theta$ (9), whose Tangent manifold $\mathbf{L}_\theta$ (35) is the submanifold $\mathbf{L}_H \subset$ Grass(p, q) (64). The reconstruction mapping g is constructed to provide the relations $\mathbf{X} \approx \mathbf{X}_\theta$ (12) and $\mathbf{L} \approx \mathbf{L}_\theta$ (36).

### *5.4.1 The nearness measures in the Coordinate space*

Let the dataset
$$\mathbf{Y}_n = \{y_1, y_2, \ldots, y_n\} = h(\mathbf{X}_n) = \{h(X_1), h(X_2), \ldots, h(X_n)\} \tag{76}$$
consists of the values of the mapping h(X) (73) at the sample points. Denote the mapping $h^{-1}(y_i) = X_i$, $i = 1, 2, \ldots, n$, defined on the embeddings $y \in \mathbf{Y}_n$.

Let $y = h(X) \in \mathbf{Y}_\theta$ be an arbitrary point from the Estimated Coordinate space (10), $X' \in \mathbf{X}_n$ is a sample point near to X, and $y' = h(X')$. It is follows from (67) that
$$X - X' \approx H(X') \times (y - y'),$$
whence comes an approximate relation
$$|X - X'| \approx |v(X') \times (y - y')|. \tag{77}$$

Based on this relation, a nearness measure in the Coordinate space is constructed in the following steps 4.1 – 4.3.

**Step 4.1 [Euclidean nearness in Coordinate space].** Put
$$u_E(y) = \{y' \in \mathbf{Y}_n: (y - y')^T \times K_v(h^{-1}(y')) \times (y - y') < (\varepsilon_1)^2\},$$
where
$$K_v(X) = v^T(X) \times v(X). \tag{78}$$

For the points $y \in R^q$ and $y' \in \mathbf{Y}_n$ define
$$k_E(y, y') = k_0(y, y') \times \exp\{-\varepsilon_2 \times (y - y')^T \times K_v(h^{-1}(y')) \times (y - y')\} \tag{79}$$
as an Euclidean nearness measure between these points, here $k_0(y, y') = 1$, if $y' \in u_E(y)$, and $k_0(y, y') = 0$ otherwise.

**Step 4.2. [Tangent nearness in Coordinate space].** This step consists of a few substeps:

**4.2.1.** By applying the PCA to the set
$$U_E^*(y) = \{h^{-1}(y'): y' \in u_E(y)\} \subset \mathbf{X}_n,$$
the ordered eigenvalues $\lambda_1^*(y) \geq \lambda_2^*(y) \geq \ldots \geq \lambda_p^*(y)$ and the corresponding principal vectors are constructed. Introduce the set
$$\mathbf{Y}_g = \{y \in R^q: \lambda_q^*(y) > \varepsilon_3\}.$$

**4.2.2.** For $y \in \mathbf{Y}_g$, define both the p×q matrix $q_{PCA}(y) \in$ Stief(p, q) whose columns consist of the first q principal vectors, and the linear space
$$L^*(y) = \text{Span}(q_{PCA}(y)) \in \text{Grass}(p, q).$$

**4.2.3.** For the points $y \in \mathbf{Y}_g$ and $y' \in \mathbf{Y}_n$, construct the q×q matrix
$$s(y, y') = (q_{PCA}(y))^T \times Q_{PCA}(h^{-1}(y')).$$
Using the Binet-Cauchy kernel on Grass(p, q), introduce a tangent nearness measure between the points $y \in \mathbf{Y}_g$ and $y' \in \mathbf{Y}_n$ as Grassmann kernel
$$k_G(y, y') = \text{Det}^2(s(y, y')). \tag{80}$$

**Step 4.3. [Aggregate nearness measure].** For the points $y \in \mathbf{Y}_g$ and $y' \in \mathbf{Y}_n$, define the aggregate nearness measure in the Estimated Coordinate space by
$$k(y, y') = k_E(y, y') \times k_G(y, y'). \tag{81}$$

**Note.** It follows from (77) and the introduced definitions (79) - (81) that the approximate equality
$$k(h(X), h(X')) \approx K(X, X').$$
is hold for the near points $X \in \mathbf{X}_h$ and $X' \in \mathbf{X}_n$.

*5.4.2 Reconstruction mapping*

Based on the conditions (30), (59) and (65), construct a matrix G(y) for y ∈ $\mathbf{Y}_g$ satisfying the condition Span(G(y)) = Span($q_{PCA}$(y)) which minimizes the quadratic form

$$\Delta_G(G(y)) = \frac{1}{2}\sum_{j=1}^{n} k(y, y_j) \times \|G(y) - H(X_j)\|_F^2.$$

A solution of this optimization problem can be obtained in an explicit form which defines the following step.

**Step4.4.** For an arbitrary point y ∈ $\mathbf{Y}_g$, define the p×q matrix

$$G(y) = q_{PCA}(y) \times (q_{PCA}(y))^T \times \sum_{j=1}^{n} k^*(y, y_j) \times H(X_j), \qquad (82)$$

which minimizes $\Delta_G(G(y))$, where

$$k^*(y, y_j) = \frac{k(y, y_j)}{\sum_{i=1}^{n} k(y, y_i)}, j = 1, 2, \ldots, n.$$

For the near points y ∈ $\mathbf{Y}_g$ and y′ ∈ $\mathbf{Y}_n$, the approximate relations (66) can be written in the form

$$g(y') - g(y) \approx G(y) \times (y' - y). \qquad (83)$$

Taking into account the relanion (8), construct a vector g(y) ∈ $R^p$ for an arbitrary point y ∈ $\mathbf{Y}_g$ by minimizing the quadratic form

$$\Delta_g(g(y)) = \frac{1}{2}\sum_{j=1}^{n} k(y, y_j) \times \|g(y) - X_j - G(y) \times (y - y_j)\|^2. \qquad (84)$$

A solution of this optimization problem can be obtained in an explicit form which defines the following step.

**Step4.5.** For y ∈ $\mathbf{Y}_g$, define the vector g(y) ∈ $R^p$

$$g(y) = g_{KNR}(y) + G(y) \times (y - \sum_{j=1}^{n} k^*(y, y_j) \times y_j), \qquad (85)$$

which minimizes $\Delta_g(g(y))$ (84), where

$$g_{KNR}(y) = \sum_{j=1}^{n} k^*(y, y_j) \times X_j. \qquad (86)$$

**Note.** It follows from (85), (86) that $J_g(y) = G(y)$ thus the relation (30) is satisfied for the points y ∈ $\mathbf{Y}_g$.

**Note.** For the near points y ∈ $\mathbf{Y}_g$ and y′ ∈ $\mathbf{Y}_n$, the approximate relations (66), (83) can be rewritten in equivalent form

$$g(y) - g(y') \approx H(h^{-1}(y')) \times (y - y'),$$

and the mimization of the quadratic form (83) can by replaced by minimizing the 'equivalent' quadratic form

$$\Delta^*(g(y)) = \frac{1}{2}\sum_{j=1}^{n} k(y, y_j) \times \|g(y) - X_j - H(X_j) \times (y - y_j)\|^2$$

with the 'equivalent' solution

$$g(y) = g_{KNR}(y) + \sum_{j=1}^{n} k^*(y, y_j) \times H(X_j) \times (y - y_j), \qquad (87)$$

see (85), (86).

*5.5 Properties of the GSE*

1) If the sample size n tends to infinity together with a tending the thresholds ε to 0 with an appropriate convergence rate then all the points from the sets **X** and $\mathbf{Y}_\theta$ will fall into the sets $\mathbf{X}_h$ and $\mathbf{Y}_g$, respectively.

2) The reconstruction error (X - $r_\theta$(X)) can be represented in the form

$$X - r_\theta(X) \approx \pi^\perp(X) \times (X - \sum_{j=1}^{n} K^*(X, X_j) \times X_j) \qquad (88)$$

which is small because of both a smallness of the second factor in the right part of (88):

$$|X - \sum_{j=1}^{n} K^*(X, X_j) \times X_j| \leq \varepsilon_1,$$

and an independent fact that the vector (X - $\sum_{j=1}^{n} K^*(X, X_j) \times X_j$) lies nearly the linear space $L_{PCA}(X)$ (38) with the projector π(X) (61):

$$X - \sum_{j=1}^{n} K^*(X, X_j) \times X_j \approx \pi(X) \times (X - \sum_{j=1}^{n} K^*(X, X_j) \times X_j),$$

so

$$\pi^\perp(X) \times (X - \sum_{j=1}^{n} K^*(X, X_j) \times X_j) \approx (\pi^\perp(X) \times \pi(X)) \times (X - \sum_{j=1}^{n} K^*(X, X_j) \times X_j) = \mathbf{0}. \quad (89)$$

3) It follows from (88) that

$$(Q_{PCA}(X))^T \times (X - r_\theta(X)) \approx (Q_{PCA}(X))^T \times \pi^\perp(X) \times (X - \sum_{j=1}^{n} K^*(X, X_j) \times X_j) = \mathbf{0},$$

so the embedding h(X) (3), (73) coincides approximately wtth the embeddinh $h_g(X)$ (14), so h(X) is the projection function onto the ED-manifold $\mathbf{X}_\theta$.

4) Consider the Jacobians $J_{g \bullet h}$ and $J_{h \bullet g}$ of the mappings $r_\theta = g \bullet h$ (11) and $h \bullet g: \mathbf{Y}_\theta \to \mathbf{Y}_\theta$, respectively. Then the following relations

$$J_{g \bullet h}(X) = \pi(X),$$
$$J_{h \bullet g}(y) = I_q, \quad (90)$$

hold true. As a consequence, the reconstruction error $(X - r_\theta(X))$ has null Jacobian, and the following relations

$$r_\theta(X') - r_\theta(X) = X' - X + o(\|X' - X\|),$$
$$h(r_\theta(X')) - h(r_\theta(X)) = h(X') - h(X) + o(\|X' - X\|),$$

hold true for the near points X, X′ ∈ **X**.

## *5.6 The orthogonal GSE*

Describe an orthogonal vertsion of GSE-algorithm, **Orthogonal GSE** (OGSE) in which the Embedding mapping h (73) has a local isometric property:

$$|X - X'| \approx |h(X) - h(X')|$$

for the near points X, X′ ∈ **X**.

**At Stage 2** in OGSE, the mapping (43) is replaced by the mapping

$$H: X \in \mathbf{X} \to H(X) \in \text{Stief}(p, q).$$

In other words, the quadratic form (48) is minimized over the matrices H having a form (46), but the normalizing condition (49) is replaced by the conditions

$$H_i^T \times H_i = I_q \text{ for all the indexes } i = 1, 2, \ldots, n,$$

whence comes that $H_i$ and $v_i$ are the orthogonal matrices, i = 1, 2, ... , n.

So, the generalized eigenvector problem (55) is replaced by a maximization of the quadratic form (53) over the orthogonal matrices $v_1, v_2, \ldots, v_n \in \text{Stief}(q, q)$, and this problem may be referred to as the Averaged Orthogonal Procrustes Problem.

The solution $v_{1,ort}, v_{2,ort}, \ldots, v_{n,ort} \in \text{Stief}(q, q)$ of this optimization problem satisfies the following relations:

$$v_{i,ort} = R\left(\sum_{j \neq i} K^*(X_i, X_j) \times S(X_i, X_j) \times v_{j,ort}\right), i = 1, 2, \ldots, n, \quad (91)$$

where the ortogonalized operator

$$R: \text{Stief-NC}(q, q) \to \text{Stief}(q, q)$$

transforms an arbitrary q×q matrix A with SVD $A = A_1 \times \Sigma \times A_2$ ($A_1$ and $A_2$ are orthogonal matrices, $\Sigma$ is diagonal matrix) to the orthogonal q×q matrix $R(A) = A_1 \times A_2$.

The relations (91) determine the orthogonal version **Step2.1ort** in OGSE for the Step2.1, in which a solution of the Averaged Orthogonal Procrustes Problem (orthogonal matrices $v_1, v_2, \ldots, v_n$) are computed by using the following iterative algorithm:

- the orthogonal matrices $R(v(X_i))$ with $v(X_i)$ (63), i = 1, 2, ... , n, are taken as the starting solution;
- the next solution's iteration is calculated as the values of the right parts of (91) computed for the solution obtained at the previous iteration.

The **Step2.2ort** in OGSE calculates the matrix H(X) by the formula (46) with

$$v(X) = v_{ort}(X) = R\left(\sum_j K^*(X, X_j) \times S(X, X_j) \times v_{j,ort}\right). \quad (92)$$

At **Stage 3** in OGSE, the least squares equations (71) for the vector set $\mathbf{h}_n$ (68) and the solution h(X) (73) can be written as

$$h_i - \sum_{j=1}^{n} K^*(X_i, X_j) \times h_j = \sum_{i=1}^{n} K^*(X_i, X_j) \times \tfrac{1}{2}\left(H(X_i) + H(X_j)\right)^T \times (X_i - X_j) \equiv$$

$$\equiv H^T(X_i) \times \left(X_i - \sum_{j=1}^{n} K^*(X_i, X_j) \times X_j\right) + \xi_i, \ i = 1, 2, \ldots, n, \quad (93)$$

and

$$h(X) = h_{KNR}(X) + H^T(X) \times \left(X - \sum_{j=1}^{n} K^*(X, X_j) \times X_j\right), \quad (94)$$

respectively, where $h_{KNR}(X)$ was defined in (75), and the quantities

$$\xi_i = \frac{1}{2} \times \sum_{j=1}^{n} K^*(X_i, X_j) \times \left\{\left(H(X_j) - H(X_i)\right)^T \times (X_j - X_i)\right\}, i = 1, 2, \ldots, n, \quad (95)$$

are small.

Because of a smallness of the quantities (95), the Steps 3.1 and 3.2 are replaced in the OGSE by the Step3.1ort and Step3.2ort, respectively.

**Step3.1ort.** Compute the vector set $\mathbf{h}_n$ (68) as a solution of the linear least squares equations with $i^{th}$ equation

$$h_i - \sum_{j=1}^{n} K^*(X_i, X_j) \times h_j = H^T(X_i) \times \left(X_i - \sum_{j=1}^{n} K^*(X_i, X_j) \times X_j\right), \quad (96)$$

$i = 1, 2, \ldots, n$, complemented by the normalizing equation (70).

**Step3.2ort.** For an arbitrary point $X \in \mathbf{X}_h$, the embedding $h(X)$ is calculated by the formulae (94), (75), where the vector set $\mathbf{h}_n$ in (75) was calculated in the Step3.1ort.

**Note.** Due an orthogonality of the matrix $v(X)$ (92) in the relations (77), the Embedding mapping $h$ (94), (75) in OGSE has a local isometric property.

At **Stage 4** in OGSE, the **Step4.1ort** differs from the Step4.1 in GSE only by replacing the $q \times q$ matrix $K_v(X)$ (78) by the identity matrix $I_q$; the Steps 4.2 and 4.3 are not changed. In the **Step4.4ort** in OGSE, the matrix $G(y)$ is computed by formula

$$G(y) = G_{ort}(y) = q_{PCA} \times R\left(q_{PCA}^T(y) \times \sum_{j=1}^{n} k^*(y, y_j) \times H(X_j)\right) =$$
$$= q_{PCA} \times R\left(\sum_{j=1}^{n} k^*(y, y_j) \times s(y, y_j) \times v_{ort}(X_j)\right). \quad (97)$$

In the **Step4.5ort**, the reconstruction function $g(y)$ is computed by the formula (85) or (87) with using the matrices $G(y)$ or $H(X)$ calculated in the Steps4.4ort or Step2.2ort, respectively.

## 6  Numerical experiments

The GSE-solution $\vartheta = (h, g, G) = (\theta, G)$ gives also a new solution $\theta = (h, g)$ for the ML. To compare the GSE solution with the known methods, the numerical experiments were performed[12]. In the experiments, OGSE-algorithm was compared with the LLE, Conformal Eigenmaps; HLLE (Hessian LLE), ISOMAP, Landmark ISOMAP and LTSA.

Two artificial nonlinear Data manifold were used in the experiments: SwissRoll (Fig. 1(a)) and Spiral in $R^3$ (Fig. 5(a)). The training datasets were sampled randomly from these manifolds, and all the algorithms were applied to the same training samples to construct Embedding and Reconstruction mappings (LLE Reconstruction was used for the algorithms without own reconstruction mapping).

After that, the independent test datasets were sampled randomly from the manifolds, and the constructed Embedding and Reconstruction mappings were applied to test points. The averaged reconstruction errors were calculated for each algorithm based on the test datasets. The results are illustrated by the following Figures.

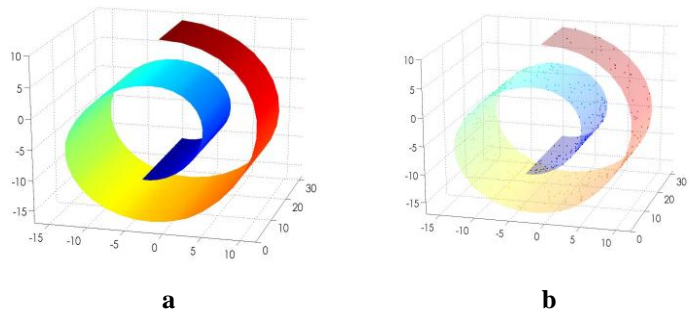

**a** **b**

Figure 1. SwissRoll manifold (a) and Training Dataset (b) consisting of $n_{train}$ = 450 points

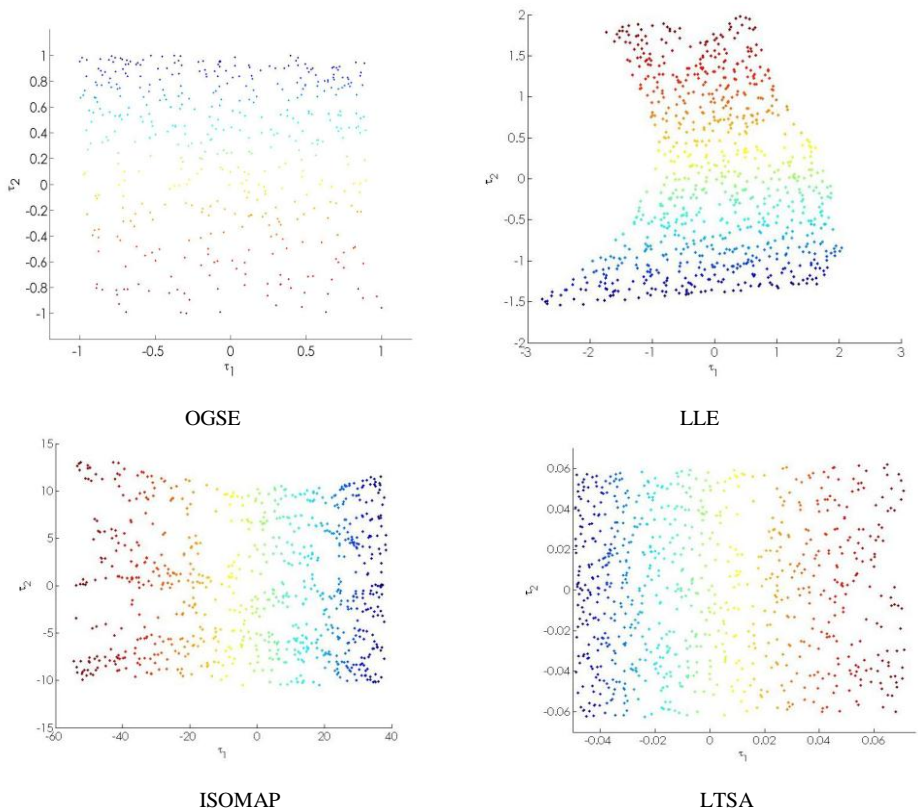

OGSE  LLE

ISOMAP  LTSA

Figure 2. Embedding of SwissRoll Training dataset by various methods

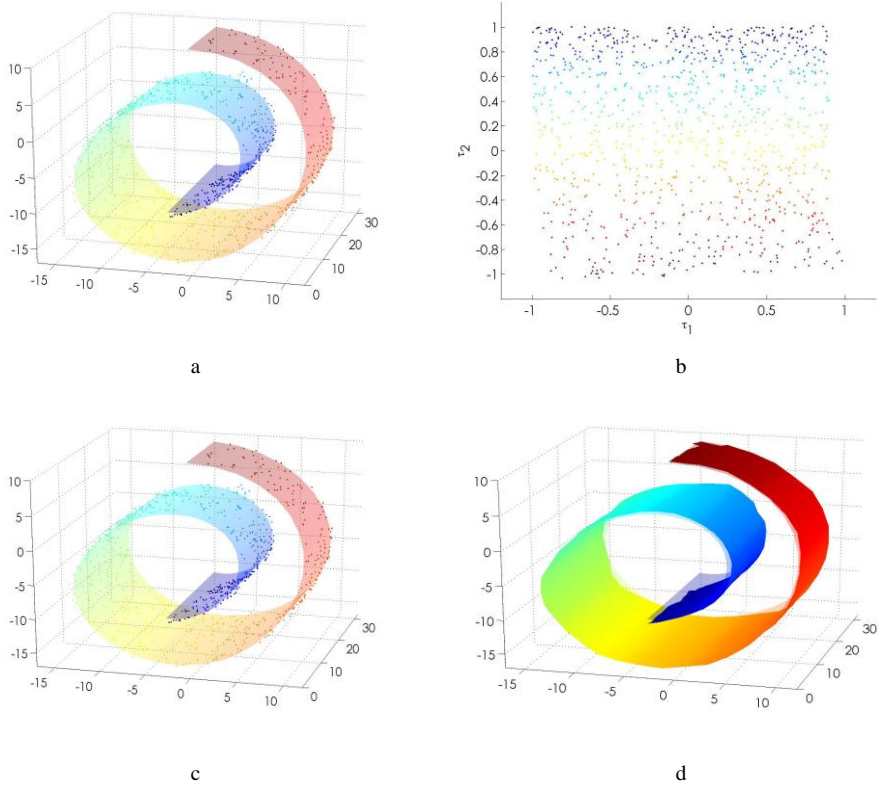

Figure 3. Results for the OGSE algorithms: (a) Test SwissRoll dataset;
(b) Embedding of Test SwissRoll dataset; (c) Reconstruction of Test SwissRoll dataset;
(d) Solid reconstruction (based on huge uniform test grid)

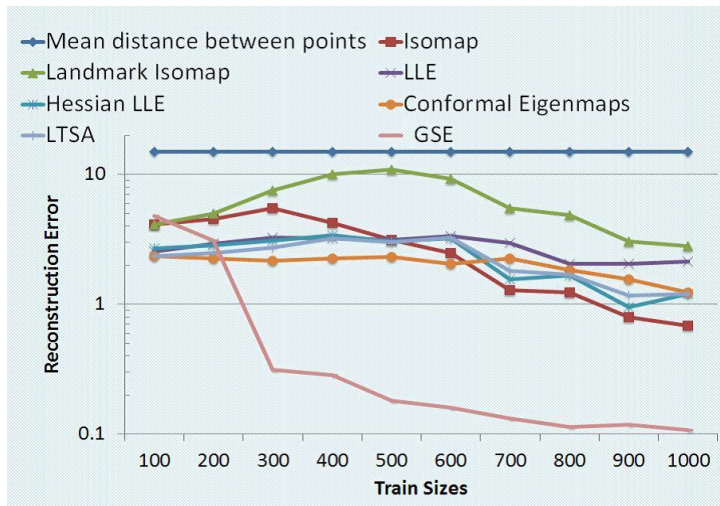

Figure 4. Mean SwissRoll Reconstruction errors for various sizes of Training dataset

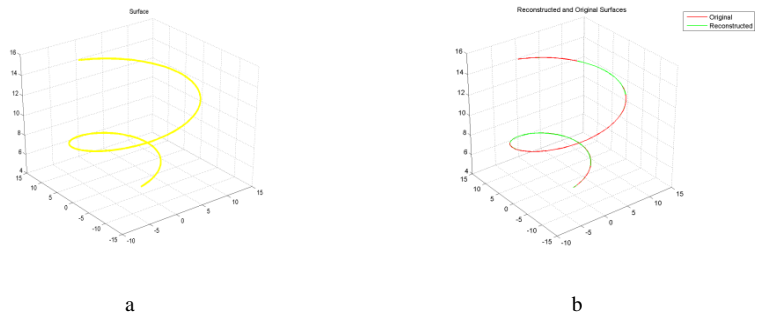

a　　　　　　　　　　　　　　　　　　b

Figure 5 Spiral manifold (a); Original and Solid Reconstructed Spiral manifold (b)

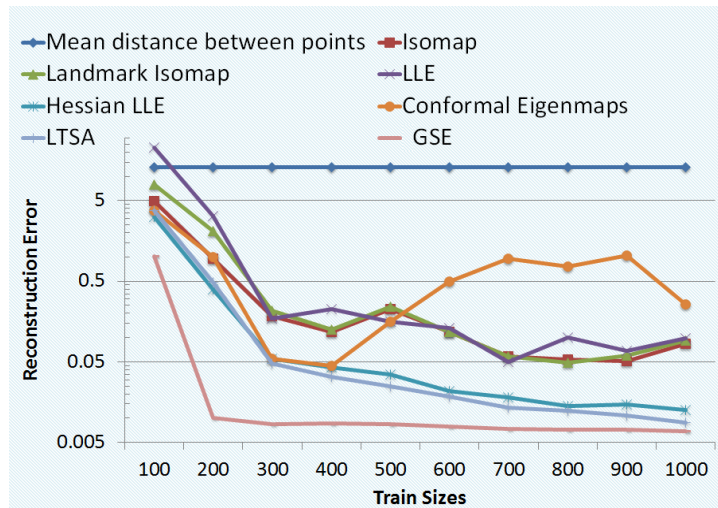

Figure 6. Mean Spiral Reconstruction errors for various sizes of Training dataset

The numerical experiments performed for other artificial manifolds demonstrate also a good quality of the GSE-algorithm.